\documentclass[a4paper,fleqn,final]{cas-dc}

\usepackage[numbers]{natbib}

\def\RAISELPBF{\textsc{\lowercase{RAISE-LPBF}}\xspace}
\def\RAISELPBFLaser{\textsc{\lowercase{RAISE-LPBF-Laser}\xspace}}

\begin{document}
\let\WriteBookmarks\relax
\def\floatpagepagefraction{1}
\def\textpagefraction{.001}

\shorttitle{Published in Additive Manufacturing Letters}

\shortauthors{C. Blanc, A. Ahar, K. De Grave}

\title [mode = title]{Reference dataset and benchmark for reconstructing laser parameters from on-axis video in powder bed fusion of bulk stainless steel}

\author[1]{Cyril Blanc}[orcid=0000-0003-3271-2398]
\credit{Methodology, Experimentation, Software, Writing}

\author[1]{Ayyoub Ahar}[orcid=0000-0001-8277-7013]
\credit{Writing, Validation}

\author[1]{Kurt De~Grave}[orcid=0000-0001-9116-6986]
\cormark[1]
\ead{kurt.degrave@flandersmake.be}
\credit{Conceptualization of this study, Experimental design, Methodology, Writing}

\cortext[cor1]{Corresponding author}

\address[1]{organization={Flanders Make vzw},
    addressline={Oude Diestersebaan 133},
    city={Lommel},
    postcode={3920},
    country={Belgium}}

\tnotetext[1]{Published in Additive Manufacturing Letters, Volume 7, December 2023, DOI: 10.1016/j.addlet.2023.100161}

\begin{abstract}
We present \RAISELPBF, a large dataset on the effect of laser power and laser dot speed in powder bed fusion (LPBF) of 316L stainless steel bulk material, monitored by on-axis 20k FPS video. 
Both process parameters are independently sampled for each scan line from a continuous distribution, so interactions of different parameter choices can be investigated.
The data can be used to derive statistical properties of LPBF, as well as to build anomaly detectors.  We provide example source code for loading the data, baseline machine learning models and results, and a public benchmark to evaluate predictive models.
\end{abstract}

\begin{highlights}
\item Comprehensive parameter sweep of laser power and laser dot speed in stainless steel bulk material printing
\item Process monitoring by high-speed on-axis video
\item Computer vision for laser parameter reconstruction
\item Public dataset and benchmark
\end{highlights}


\maketitle

\section{Introduction}
Powder Bed Fusion (PBF) is the most common Additive Manufacturing (AM) method where the powder of the chosen material (e.g., nylon, or various types of metal powder) is heated to construct the product. The most common heat sources are laser beam (LPBF, sometimes written L-PBF) or electron beam. The laser based heating process may result in sintering powder in a process called selective laser sintering or completely melt (typically metal) powder in selective laser melting. In the latter method, a moving laser beam follows a pre-planned pattern derived from the CAD model to melt and add the powder to previous printed layers. 

Successful production of parts with LPBF is a delicate process which depends on several factors, particularly the laser beam parameters like power and velocity. Various micro-structural anomalies and production defects may occur during the printing that degrade the structural integrity and quality of the build \cite{Mostafaie2022}. Among others, tiny voids and pores inside the product may be created mainly due to unstable printing conditions and variations in laser power and speed \cite{booth22}. Current solutions to reduce such defects are mostly heuristic, involving costly post-production testing via both destructive and non-destructive approaches which thereafter will guide the readjustment of the laser beam parameters. However, it requires several repeated printings and thus significant increase in product scrap-rate, which imposes a higher cost to the final product.

A better solution for reducing the defective prints is an automated monitoring system embedded in a low-latency feedback control loop that enables on-the-fly supervision of the printing process. In such a solution, the monitoring system observes the meltpool and monitors a set of print parameters to predict whether the current trajectory of the printing is expected to end up in generating pores. Consequently, the control unit may generate corrective control measures to prevent the generation of defects in the first place, or compensate for it by remelting the faulty region. 

The very fast movements of the melting laser, i.e., laser dot speeds up to $1500mm/s$, impose strict time constraints on the entire cycle of monitoring-defect prediction-control feedback. This challenge further intensifies when considering the fact that accurate steering of the laser beam requires detailed information about the spatial location of the detected printing events or anomalies. Therefore, simple optical sensors like photodiodes are not the best option \cite{craeghs2011online,bisht2018correlation}. Instead, high frame-rate 2D cameras have been favoured recently due to providing a better balance between temporal and spatial resolution of their output. To clarify the typical computational load for this process, let us approximate printing one layer of an object with a $6$x$6mm$ square section. To print a layer of this object with $\approx60$ laser scan lines and a fixed laser speed of $900mm/s$, it will take $\approx6.7ms$ to print one line and $0.4s$ for a full layer. A typical high-speed camera of $20k$ FPS will produce more than $133$ images of $\geq 10k$ pixels per print line which means that to have at least one control feedback signal after printing each $10$ lines, the full monitoring-defect prediction pipeline should process $\approx1334$ frames in $\approx67ms$. This time slot includes capturing and transferring images into the memory of a computational device, feature extraction, and finally defect prediction. Even with good hardware integration and optimization solutions, the allocated time slot for the prediction model currently remains below anything but simple regression techniques with consequent limited accuracy\footnote{This limitation assumes full analysis of all events. Downsampling in time and/or cropping can drastically reduce the number of pixels to analyse, at the expense of not necessarily observing all events in detail.}. Nevertheless, deep learning and improving prediction models for achieving better prediction performance can be foreseen in the future. Additionally, by easing the time constraints for the feedback loop (e.g., providing control feedback once per printing layer) utilizing such solutions with higher computational complexity can already be envisaged. 

Similar to other use cases of deep learning, the first step is to provide an open-access comprehensive annotated dataset for training those artificial intelligence (AI) models. Additionally, a unified test-bed plus a set of performance assessment metrics has to be defined for fair comparison and benchmarking proposed solutions. To do so, in this paper we introduce our LPBF defect detection test-bed, Makebench, which is publicly available at \url{https://www.makebench.eu}. Our annotated reference dataset \RAISELPBF is available for download via the same web portal, except for the labels of the test set, which Makebench uses to benchmark new model submissions from the public.

In the next section we will provide a review of the existing related research that used deep learning for the purpose of analyzing videos captured from an LPBF process. In section~\ref{LPBFsetup}, we provide details of our LPBF printer where the data has been produced and its monitoring setup. Next in section~\ref{expDesign} we describe the printed objects comprising the \RAISELPBF dataset and explain the objectives behind their design.  
In section~\ref{stats}, we describe the dataset. In section~\ref{benchmark}, we present the results for a set of AI models which serve as baselines for our benchmark.  Finally, section~\ref{conclusion} concludes this paper.

\section{Related work}

In-situ monitoring of LPBF processes to detect or understand defect formation has been investigated based on different monitoring sensors covering a wide range of the light spectrum, including visible light, infrared, and X-ray, as well as acoustic emissions~\cite{smoqi2022monitoring}. Also pyrometers have been used~\cite{chivel2010line} to
track the surface temperature profile variations~\cite{pavlov2010pyrometric}. Laser induced breakdown spectroscopy (LIBS) has been proposed for in-situ quantitative elemental analysis and failure detection~\cite{lednev2019situ}.  It has long been demonstrated that print quality can be improved by feedback control, based on either a photodiode or a CMOS camera \cite{kruth2007feedback}.  Photodiodes provide a choice of wavelengths and a high sampling rate \cite{furumoto2009study}.  High frame-rate 2D cameras either in the visual \cite{craeghs2011online,clijsters2014situ,thanki2023off,booth22} or the infrared part of the spectrum \cite{khanzadeh2018porosity,ozel2018process,estalaki2022predicting, ren2023machine} provide a better balance between temporal and spatial resolution of their output.  The high data rate that such cameras produce is sometimes fed into a field-programmable gate array (FPGA) \cite{craeghs2011online} or GPU for parallel processing. The literature around AM process in-situ monitoring is rich and beyond the scope of this paper to review; the interested reader is referred to review papers like~\cite{mccann2021situ, grasso2021situ,Mostafaie2022} for a detailed classification of different sensing methodologies utilized in AM processes.

Parallel to the introduction of several in-situ monitoring methods, machine learning (ML) approaches have been widely utilized to derive useful patterns and classification of defects. Extensive reviews of past research for applying ML on various monitoring data of AM processes are available for example in~\cite{qi2019review_MLforAM,mahmoud2021_review_MLforLPBF, goh2021review_MLforAM,sahar2023anomaly_review_MLforLPBF}. Some of the conventional ML approaches, among several others, include multi-linear principal component analysis on thermal images for print anomaly detection~\cite{khanzadeh2018dual}, defect detection using a linear support vector machine (SVM) on high resolution digital images~\cite{gobert2018application}, the combination of Gaussian mixture models with randomized singular value decomposition features on photodiode data~\cite{okaro2018automatic}, and using an SVM for defect classification of acoustic signals~\cite{ye2018defects}. A comparative study of the performance of six methods for a binary defect classification task was provided in~\cite{estalaki2022predicting}, comparing K-nearest neighbors, random forests, decision trees, multi-layer perceptron, logistic regression, and AdaBoost. 
Several groups have reported using convolutional neural networks (CNN) for process anomaly and defect detection from IR thermal images or regular visual light images of the meltpool~\cite{caggiano2019machine, westphal2021machine,ansari2022convolutional,baumgartl2020deep,kwon2020deep}. A spectral CNN-based detection model also is implemented to process acoustic signals~\cite{shevchik2018acoustic}. In~\cite{ye2018defect} a deep belief network is implemented for defect detection based on the gathered acoustic signals as well. 
A hybrid deep learning model consisting of CNNs and Long-Short term
memory (LSTM) has been proposed to process multiple sensory input data including three photodiodes and acoustic sensor with varied number of data points~\cite{pandiyan2022deep}. 

There is, however, a lack of publicly available LPBF datasets for training computer vision.  As of early 2019, no public datasets with varying laser parameters were available \cite{bodi2019}.  For directly regressing mass density (or the lack of it, i.e., pores) from process parameters, a relatively small amount of data points are available \cite{barrionuevo2021comparative}.

A closely related work is the Additive Manufacturing Benchmark Test Series (AM-BENCH) which is a biennial simulation challenge started from 2018 and the latest version has released on May 2022~\cite{AM_Bench_Modelling2022}. The challenge is designed for the simulation experts to improve modeling of the LPBF process. A large database of the thermographic monitoring for sample 3D prints with nickel-based superalloy IN718 is provided which includes in-situ recording of the full build-plate using 8k FPS near-infrared camera and build plate temperature recording using two thermocouples mounted under the build plate and another one above within the build chamber. It should be noted that all test objects in this data set are printed with fixed nominal laser settings and thus not directly comparable with our dataset aimed for the ML training purposes.

\section{LPBF setup}\label{LPBFsetup}
Our LPBF setup consists of a 3D Systems ProX DMP320. The printer room is maintained at an ambient temperature of 20°C. The powder is 316L stainless steel with a particle size range between 20--50 $\mu m$. The build chamber is vacuum-cycled and filled with argon at $130mbar$ with an oxygen concentration below $100ppm$ before commencing the LPBF process. The pressure and purity is maintained during the print thanks to a constant gas flow in y-positive direction as defined in Figure~\ref{FIG:floorplan}. The printer employs a $500W$ power-adjustable IPG ytterbium fiber laser with a wavelength of $1064nm$. Each of these print layers are created by several laser scan lines. For the specimens in this work, layer thickness was set at $30\mu m$, with a laser focal spot of $75\mu m$ and hatching distance $100\mu m$. The chosen hatching strategy is perimeter scanning as described in \cite{dewidar_processing_2003}: perimeter first, then back and forth bulk lines. The hatching pattern is rotating of 67° counter-clockwise increments every layer. An example layer is shown in Figure~\ref{FIG:slice}.


The printing process is monitored with an on-axis Mikro\-tron EoSens\textsuperscript{\textregistered} 3.0MCX5 camera. It captures the LPBF process at $20k$ FPS in 8-bit grayscale 128x128 frames with $30 \mu s$ exposure time. Its on-axis nature keeps the laser dot at a fixed position in the frame, independent from its movements over the printer bed. This is achieved thanks to a SCANLAB dichroic beam splitter that decouples light reflected from the printer chamber on one side and the laser beam on the other side. Thereby, the light is decoupled from the beam path and can then be directed to the camera enabling consistent laser working field monitoring with minimal laser beam attenuation. Note that the laser spot, and hence the melt pool, is not perfectly centered in the image, because the camera's region of interest can only be set with limited accuracy.  The imaged area has sides of $4.11 \pm 0.08 mm$, or $32.11 \pm 0.60 \mu m$ per pixel.

Simultaneously, the printer controller's information about the laser (on/off status, position, speed, power) is recorded at 100kHz. Both data streams share a common signal for syncing purposes, the laser on/off status. The frame grabber retrieves the binary status from a direct signal wire connection to the printer and embeds this bit into every frame's metadata record. A custom algorithm aligns the video metadata and the printer control logs during post-processing.

\section{Design of Experiment}\label{expDesign}

\begin{figure}
	\centering
		\includegraphics[scale=.35]{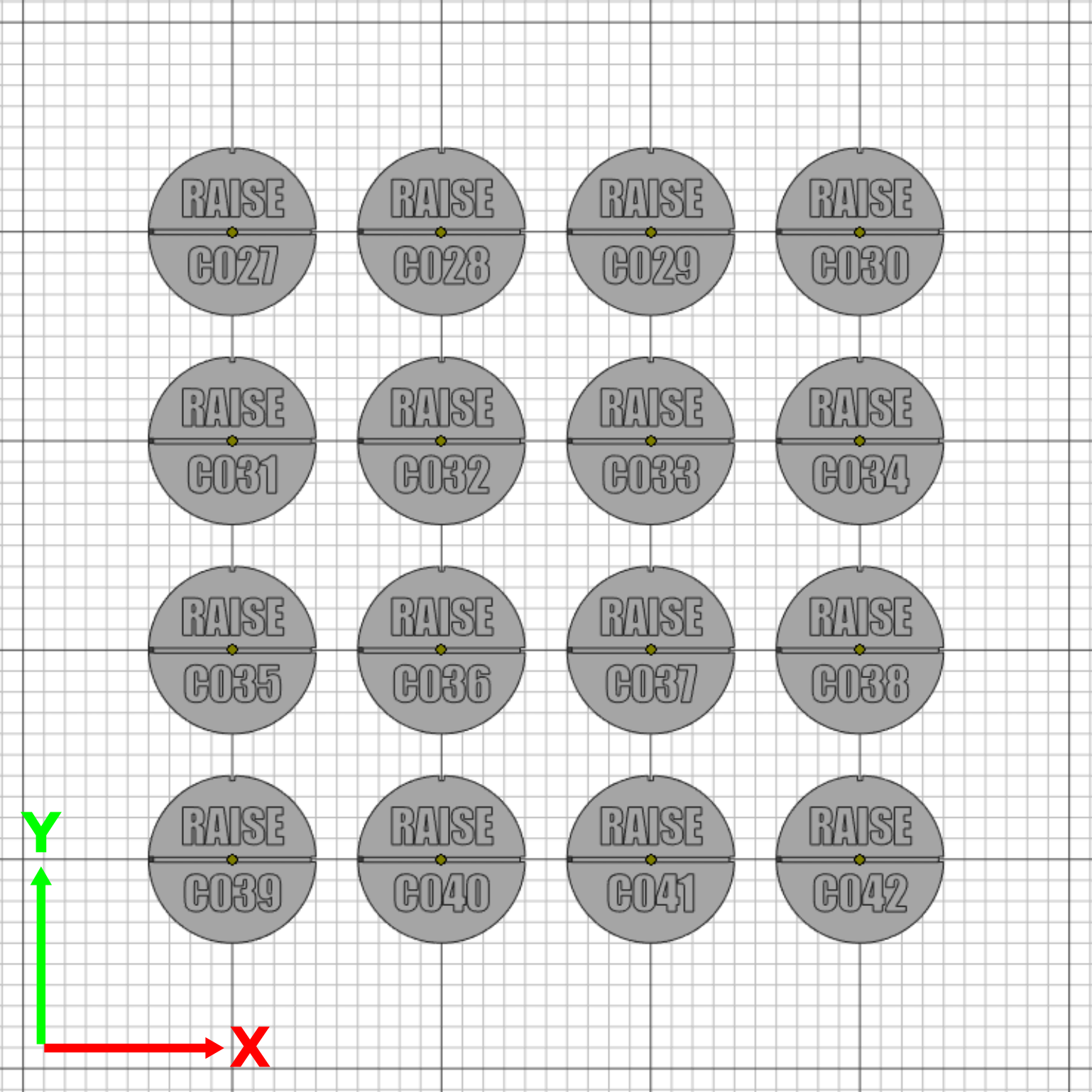}
	\caption{Top view of the objects layout on the build plate. Fine grid size is 1mm.} 
	\label{FIG:floorplan}
\end{figure}

The experiment consists of the printing of 16 individual objects at the same time, arranged in an axis-aligned 4-by-4 grid centered on the 
build plate, the occupied central part of which is depicted in Figure~\ref{FIG:floorplan}. All objects share the same 3D design, but their printing laser parameters are independently sampled, making each one unique with different induced defects. 


\begin{figure}
	\centering
		\includegraphics[scale=.18]{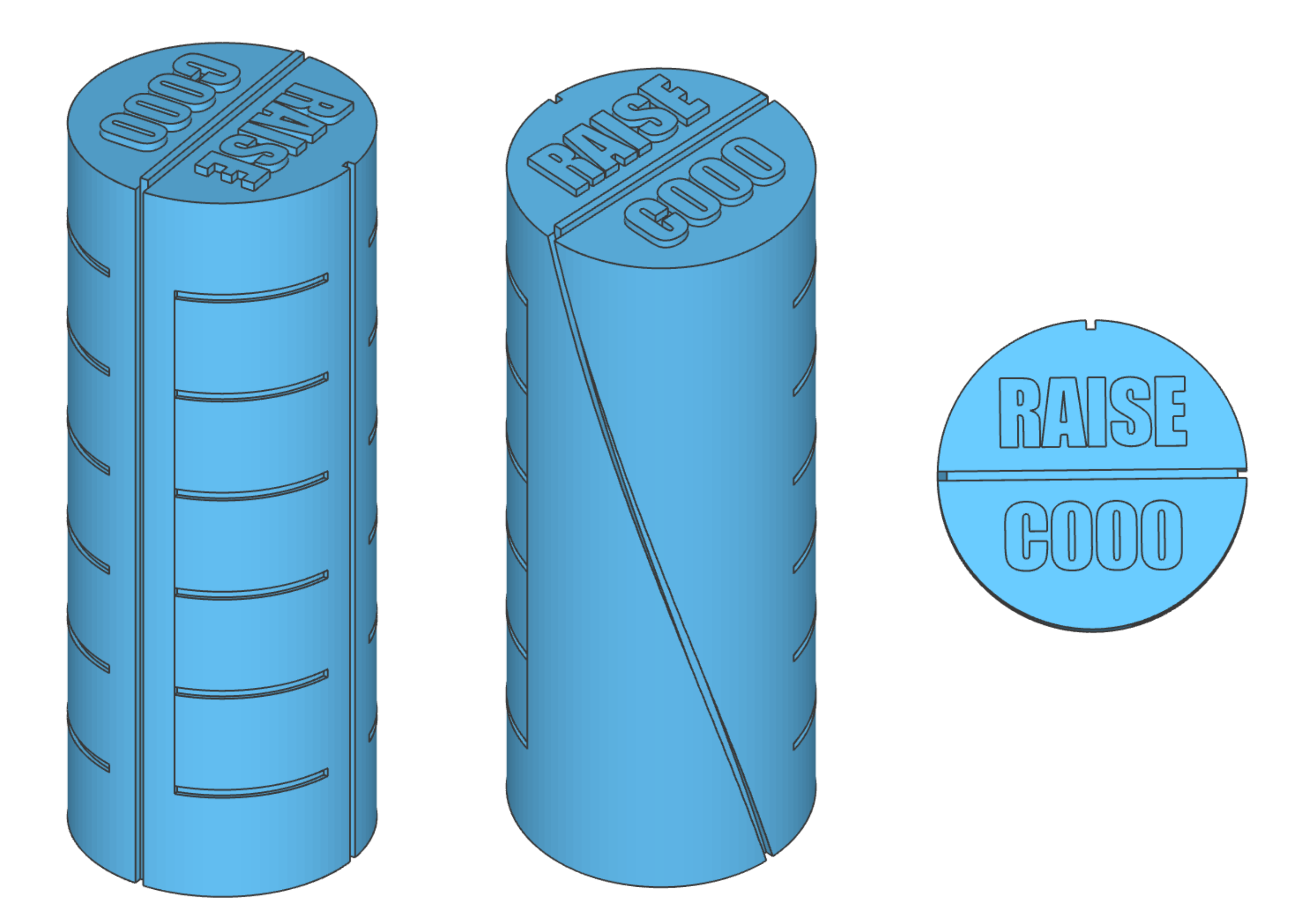}
	\caption{Renderings of the 3D design from three viewpoints.}
	\label{FIG:cad}
\end{figure}

Based on early experiments and evaluation of multiple CT scanning service providers, we selected a cylinder with 8mm diameter as our base shape to have reasonable confidence of resolving keyhole pores and lack-of-fusion zones. It seems to be the right trade-off between CT scanning and LPBF limitations, simplicity, and volumetric density of useful data.
Our designed objects (shown in Figure~\ref{FIG:cad}) are printed upright and are $21mm$ tall, including the 31 base layers and excluding the readable label on top and the support structures.  For downstream data alignment, the design must allow breaking symmetry and should contain distinguishable features.  For this purpose, horizontal, vertical, and helical indents are added as well as a human-readable unique identifier at the top. The indents are all $240\mu m$ deep inside the base cylinder shape; the horizontal ones are also $240\mu m$ high, each one is a $60^{\circ}$ arc and they are evenly spaced every $3mm$; the vertical ones take up the full height of $21mm$ and are placed $90^{\circ}$ apart; the helical indent has a $84mm$ pitch so it revolves $90^{\circ}$ for the full cylinder.


\begin{table}[width=.65\linewidth,cols=2,pos=ht]
\caption{Layers in each object, top to bottom.}\label{tbl:layers}
\begin{tabular*}{\tblwidth}{@{} RL@{} }
\toprule
Layers interval & Laser parameters\\
\midrule
$[700, 707]$ & nominal \\
$[31, 699]$ & i.i.d. sampled per line \\
$[11, 30]$ & nominal \\
$\{10\}$ & lack-of-fusion \\
$[0, 9]$ & nominal \\
\bottomrule
\end{tabular*}
\end{table}

\begin{figure}[ht]
	\centering
		\includegraphics[scale=.6]{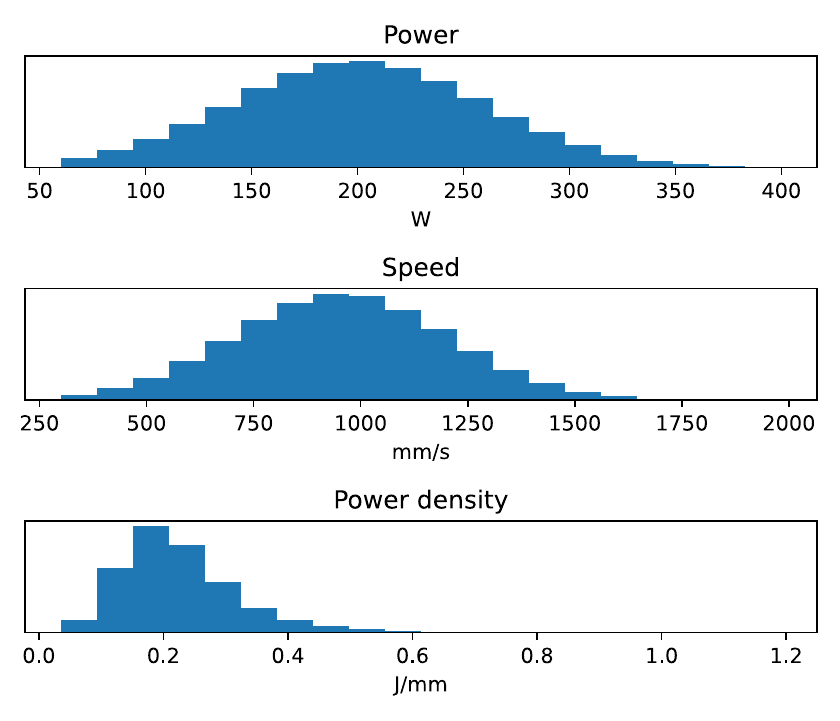}
	\caption{Histograms for laser dot speed, power, and linear power density.}
	\label{FIG:trunc_dist}
\end{figure}

\begin{table}[width=.9\linewidth,cols=5,pos=ht]
\caption{Parameters of the truncated normal distributions. Mean and standard deviation are for the original Normal distributions, before truncation.}\label{tbl:trunc_norm}
\begin{tabular*}{\tblwidth}{@{} LLLL@{} }
\toprule
 & Mean & Std. dev. & Bounds \\
\midrule
Power ($W$) & $215 \times 0.93$ & 80 & [60,400]\\
Speed ($mm/s$) & $900 \times 1.06$ & 400 & [300,2000]\\
\bottomrule
\end{tabular*}
\end{table}

\begin{figure}[ht]
	\centering
		\includegraphics[scale=.15]{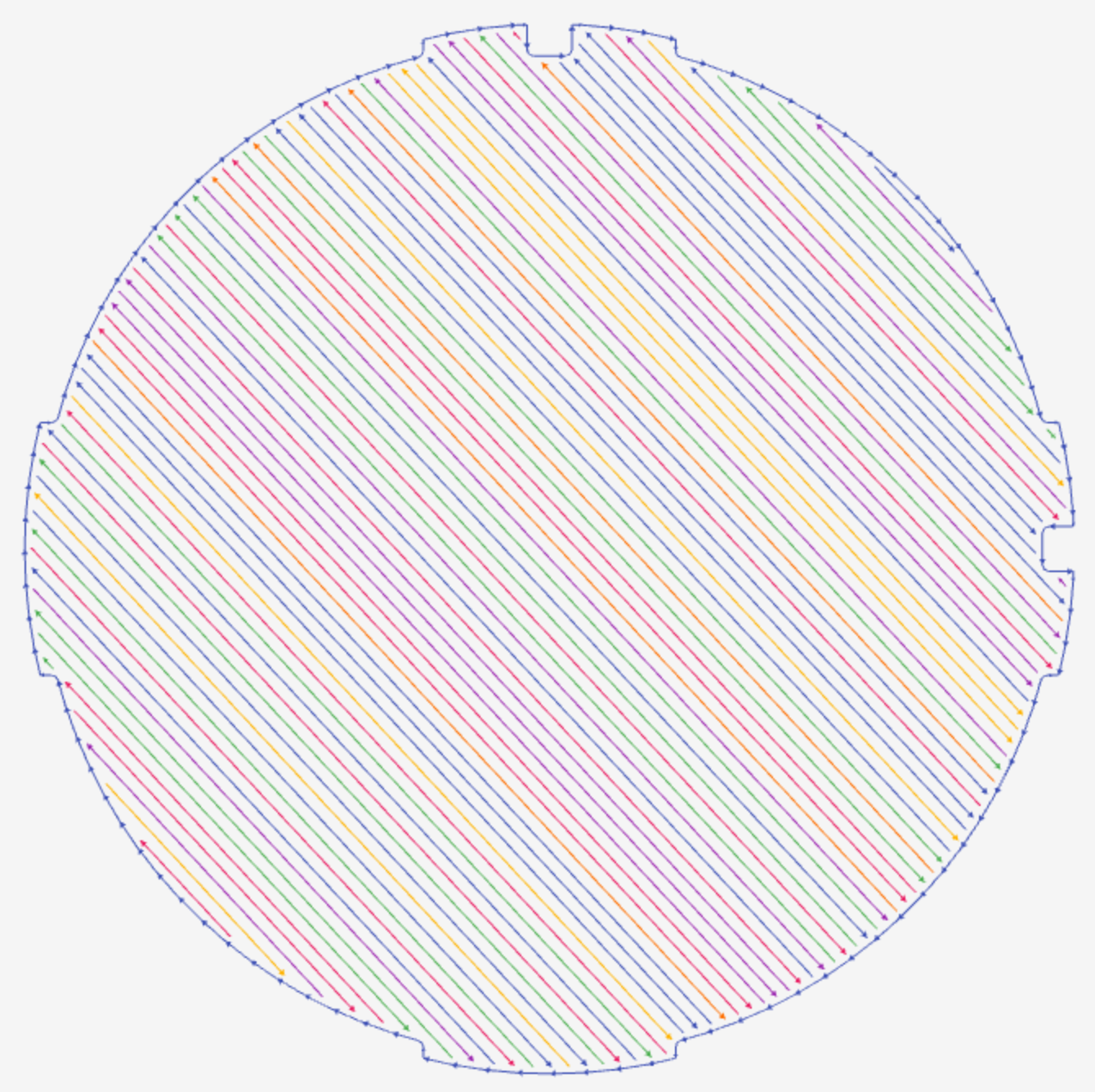}
	\caption{Example laser path over a layer for one object. Every scanline is colored arbitrarily to symbolize the independent sampling of laser speed and power.}
	\label{FIG:slice}
\end{figure}

With the objectives of introducing variation and defects for later detection and exploring laser parameter space, the speed and power of all the hatching laser scan lines (i.e., the within-object, bulk material laser vectors) between layers 31 and 699 (inclusive, see Table~\ref{tbl:layers}) are randomly i.i.d. sampled. The perimeter scanning is not modified, i.e., the laser parameters remain at $450m/s$ and $100W$ for the contours during the whole print.
We select truncated normal distributions.
The truncation is needed to satisfy the physical limits of the printer device and to avoid undesired physical effects such as burn-through in the case of extreme outliers, while still enabling a large variance. 
The underlying normal distribution is centered at manufacturer-recommended printing power ($215W$) and speed ($900mm/s$), also referred to as nominal printing settings, adjusted by a ratio so that approximately nominal linear power density (power/speed) is obtained on average\footnote{The expectation of the ratio of two independent random variables is not in general equal to the ratio of their expectations.  The introduction of variance for power and speed changes their average ratio, hence the need for the adjustment.}. Since the hatching distance is constant, linear power density is directly proportional to volumetric energy density.  Hence, the adjustment also realises nominal average volume energy density and large-scale thermal conditions, ensuring a good overall print.   Table~\ref{tbl:trunc_norm} lists the parameters of the distributions and Figure~\ref{FIG:trunc_dist} plots them. In Figure~\ref{FIG:slice}, the laser path in a sample print layer is shown. The arbitrary coloring of the scan lines represents the various laser speed values which were sampled randomly for each line (power is similarly but independently sampled). The orientation of the scan lines alternates between layers. They can be deduced from the \texttt{position} table in the training data. The test objects use the same orientation.
In the other layers, the laser parameters are not modified (as shown in Table~\ref{tbl:layers}) to create margins for the useful data. With the exception of the $10^{th}$ layer in which the bulk laser parameters are modified to intentionally create lack-of-fusion defects ($900mm/s$, $60W$). This is to help in the 3D alignment of later CT scans.

\section{Training data}\label{stats}

\begin{figure*}
	\centering
		\includegraphics[width=0.8\textwidth]{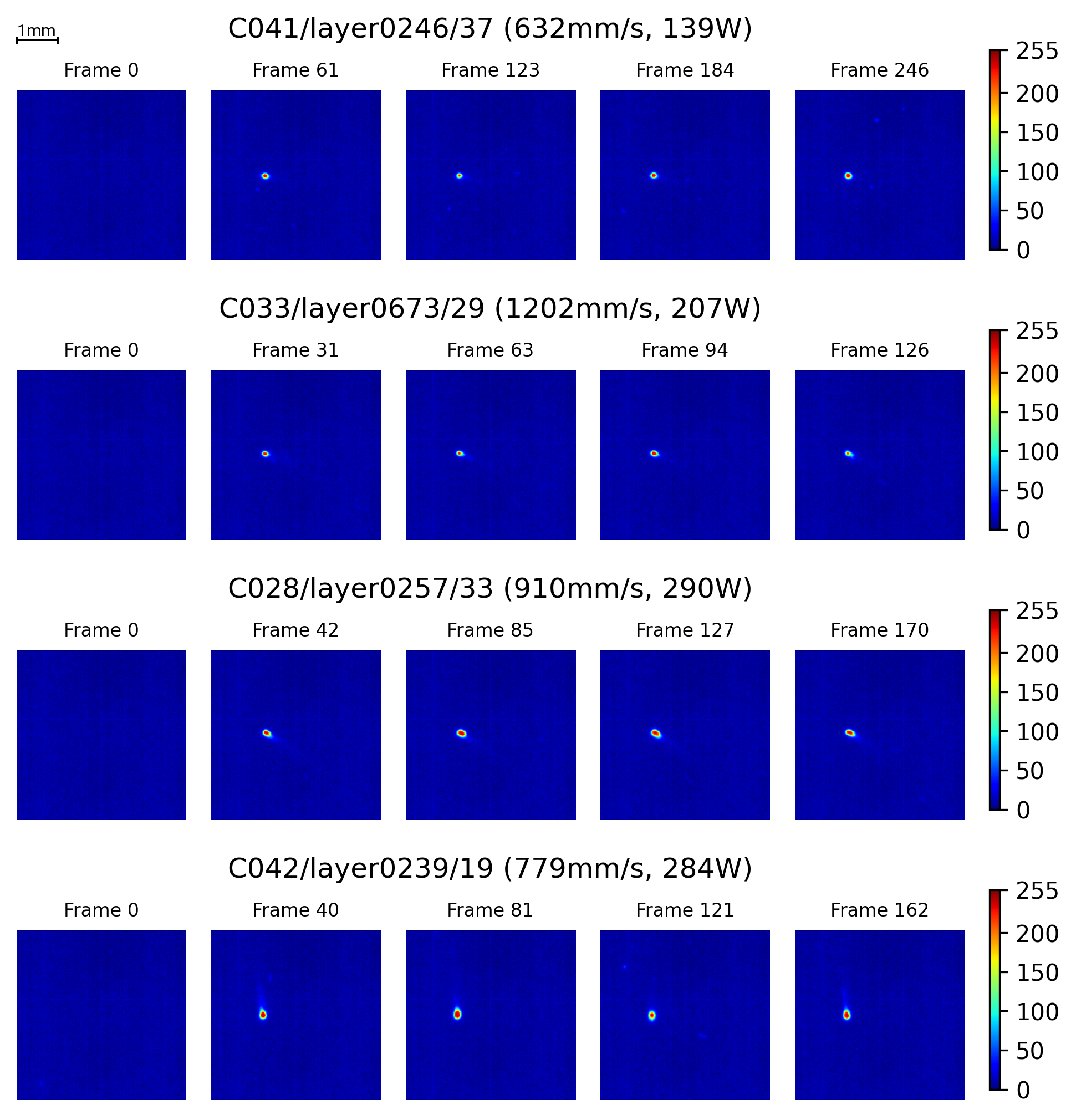}
	\caption{Example frames from random scan lines with the corresponding laser parameters.}
	\label{FIG:frames}
\end{figure*}

The training data comprises 12 cylinders.  As listed in Table~\ref{tbl:layers}, each cylinder has 669 randomized layers. The average number of scan lines per layer is $80.8$. Only bulk scanning data is considered, meaning that the contours printing phase is ignored as well as the top 8 layers that constitute the ID, and only frames with the laser active are retained in the dataset. To each line corresponds a pair of laser parameters: speed in $mm/s$ and power in $W$. While the setpoint is sampled exactly from the distributions described in Section~\ref{expDesign} and Table~\ref{tbl:trunc_norm}, the laser scan needs to be implemented on a real-world controller with finite bandwidth, which results in minor deviations from the setpoint.  The pair of parameters we provide as ground truth is the median of the measured quantities provided by the laser controller along the complete line.

\begin{figure*}[htb]
	\centering
        \includegraphics[width=0.8\textwidth]{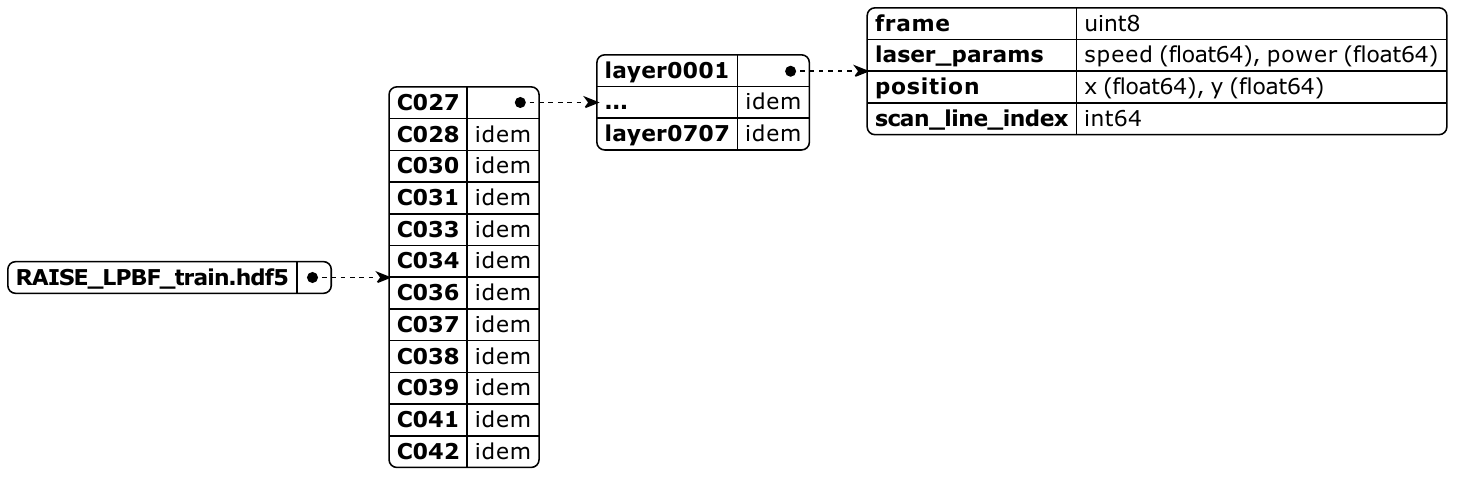}
	\caption{Structure of HDF5 dataset for the train fold. The test fold does not include the $position$ and $laser\_params$ fields.}
	\label{fig:h5train-illust}
\end{figure*}

We provide the data as files in HDF5 format, one for training and one for testing.  The size on disk of the dataset is around 1TB, using HDF5-native chunked lossless gzip compression. Frames are compressed individually, so they can be accessed randomly without excessive overhead.
The structure of the training set is illustrated in Figure~\ref{fig:h5train-illust}. The table scan\_line\_index provides for each camera frame which of the scan lines the frame belongs to. The scan lines are numbered sequentially as they are printed, restarting from 0 in each layer of each object. Table \ref{tbl:desc_data} describes in more details the number of frames and scan lines for the 8400 layers of the training dataset and Figure \ref{FIG:frames} depicts a typical sample of the data. 

\begin{table}[width=.9\linewidth,cols=3,pos=ht]
\caption{Distribution of data per layer in the training set}\label{tbl:desc_data}
\begin{tabular*}{\tblwidth}{@{}lrr@{}}
\toprule
{} &        Frames per layer &       Scan lines per layer \\
\midrule
Mean  &  11085.3 &    80.8 \\
Std. dev.   &   581.7 &    3.37 \\
Min.   &   8985 &    74 \\
Max.   &  14305 &   125 \\
Total &   93116645 &  678708 \\
\bottomrule
\end{tabular*}
\end{table}

\begin{figure}[ht]
	\centering
        \includegraphics[scale=.27]{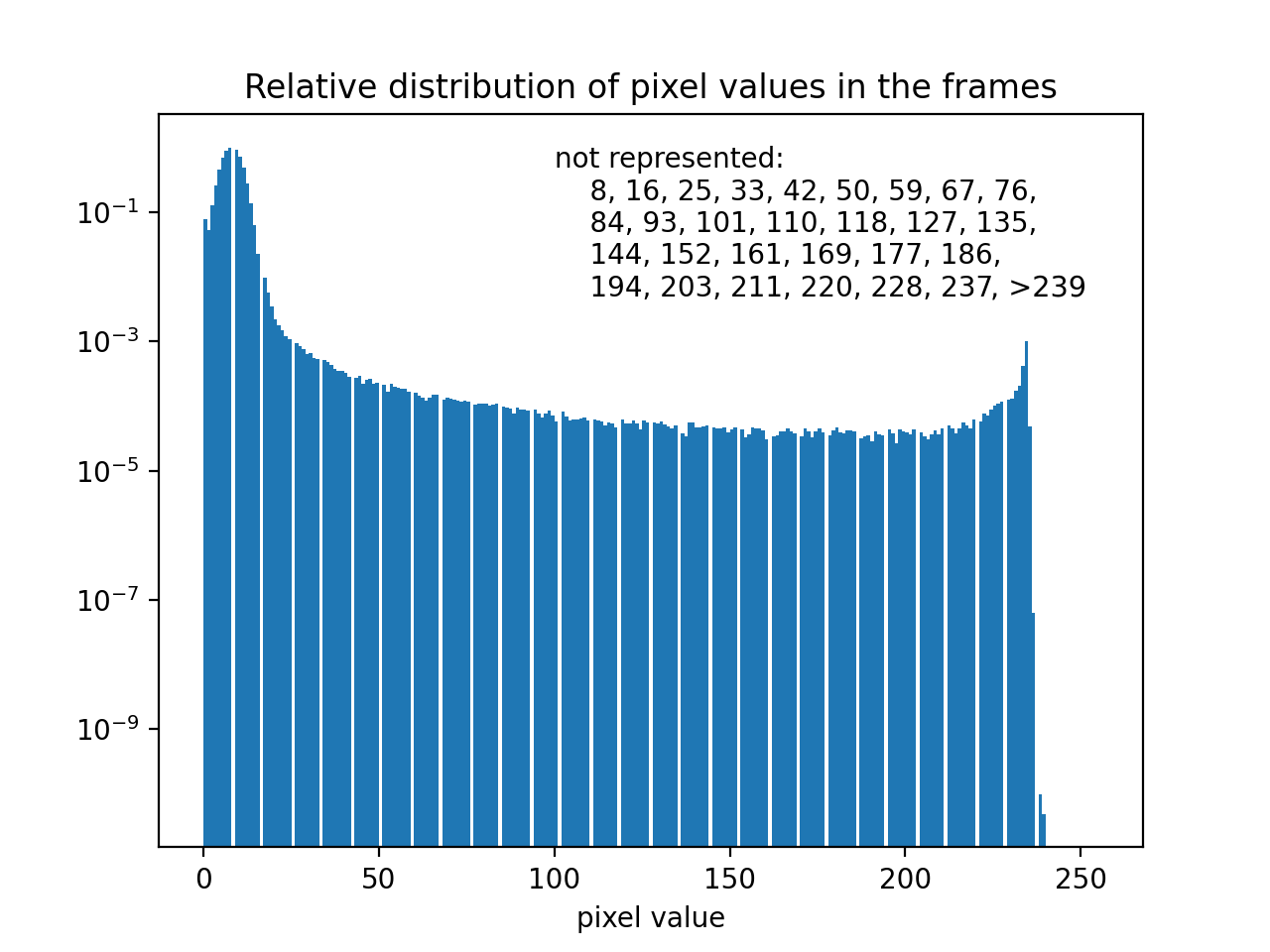}
	\caption{Probability distribution of pixel (brightness) values in all video. Some pixel values do not occur due to a camera peculiarity.}
	\label{fig:dist-brightness}
\end{figure}

As can be observed in Figures \ref{FIG:frames} and \ref{fig:dist-brightness}, the sensor data from the high-speed camera is imperfect. A constant noise pattern is present and many pixel values are never observed. These are caused by limitations of the camera and exemplifies the challenges of sensor data. We did not apply any correction to the frames and leave the application of any denoising, correction, or scaling algorithm open.

\section{Benchmark and baseline models}\label{benchmark}

We challenge the community with \RAISELPBFLaser: a public, permanent machine learning benchmark.
The task is to reconstruct the power and speed of the laser from the video input. One laser parameter tuple must be provided per scan line in the test set. The predictions are submitted in the form of a comma separated values (CSV) file as documented at \url{https://www.makebench.eu}.  We rank the models based on the root-mean-square error (RMSE) on either power, speed, or linear power density (power/speed) on the randomized layers (i.e., between 31 and 700 of the tests objects as detailed in Table \ref{tbl:layers}). Submitted models should be published eventually as source code, a research paper, or ideally both.

There is an obvious side channel for speed: the number of camera frames in a scan line provides direct information about speed, and especially so when also taking the known geometry into account.  While we could eliminate the side-channel by slicing up the test set in fixed-length chunks and shuffling them, we prefer to keep open the flexibility for the community to develop models with different window sizes, or no window size limitation at all. We just ask not to use the side channel.


\begin{table}[width=.9\linewidth,cols=4,pos=ht]
\caption{Benchmark results of baseline models}\label{tbl:baselines}
\begin{tabular*}{\tblwidth}{@{} LRRR@{} }
\toprule
Model & Speed RMSE & Power RMSE & PD RMSE \\
\midrule
3D~ResNet & 157.8 & 19.50 & 0.0520 \\
SlowFast & \textbf{67.6} & \textbf{11.12} & \textbf{0.0290} \\
MViT & 191.9 & 24.88 & 0.0623 \\
Swin3D & 205.0 & 25.34 & 0.0568 \\
\bottomrule
\end{tabular*}
\end{table}


For comparative purposes we provide results in Table~\ref{tbl:baselines} for four off-the-shelf open-source video action classification models on the described benchmark. The models were trained on a randomly sampled $70\%$ subset of the training data; the remaining data was left for validation. They were then all evaluated on the test set for the laser parameters reconstruction task. 
Video action prediction models are designed for 30 or 60 FPS video of slow actions, whereas LPBF is much faster. To obtain a prediction for a scan line, the frame sequence of the line is temporally equidistant subsampled along its full length to collect the required number of input frames for the default implementation of each model ($8$ frames for 3D~ResNet, $32$ for SlowFast, and $16$ for MViT and Swin3D).

These models can be applied, as described with a simpler model in \cite{booth22}, to perform anomaly detection by comparing the known laser parameters setpoints with model predictions. A major difference between them correlates to a high probability of anomaly. Alternatively, the features learned by the speed-power deep learning model can be reused to train a model for another task, such as pore prediction or direct anomaly classification, using much fewer labels.

The 3D-CNN-based methods 3D~ResNet \cite{hara_learning_2017} and SlowFast \cite{feichtenhofer_slowfast_2019} achieve better accuracy than their attention-based counterparts in Table~\ref{tbl:baselines}. The SlowFast model performs best by a great margin with less than half the RMSE of the second best, 3D~ResNet, for both prediction targets. Despite the recent breakthroughs of Transformers in computer vision applications, the evaluated MViT \cite{fan_multiscale_2021} and Swin3D \cite{yang2023swin3d} performed worst on the benchmark. We hypothesise that their design is less suited for the task at hand and could benefit from more input frames ($16$ vs. $32$ for SlowFast).

\section{Conclusions and future work}\label{conclusion}

At \url{https://www.makebench.eu}, the reader can find \RAISELPBF, a terabyte-sized LPBF dataset on the effect of laser power and speed. 
It can be used for laser parameter reconstruction, anomaly detection, spatter detection, and spatter prediction. 
At the same website, we challenge the machine learning community with a permanent benchmark for laser parameter reconstruction, for which we provide baseline models and results. In the future, we expect to continue to develop prediction algorithms and benchmark them on Makebench, which we hope others will do as well.  We will also extend the dataset with CT scans and a benchmark to predict porosities.

\section*{Declaration of Competing Interests}

The authors declare that they have no known competing financial
interests or personal relationships that could have appeared to influence
the work reported in this paper.

\section*{Acknowledgements}
Emma van Doren created the prototype of the website  \href{https://www.makebench.eu}{Makebench.eu} and its submission system. This article is a result of the \href{https://coe-raise.eu}{CoE RAISE project}, which has received funding from the European Union’s Horizon 2020 – Research and Innovation Framework Programme H2020-INFRAEDI-2019-1 under grant agreement no. 951733.
Part of the computational resources and services used in this work were provided by  \href{https://www.vscentrum.be/}{VSC} (Flemish Supercomputer Center), funded by the Research Foundation Flanders (FWO) and the Flemish Government – department EWI, and another part by the J\"ulich Supercomputing Center (JSC) --- including the data distribution service.

\printcredits

\bibliographystyle{model1-num-names}

\bibliography{raise}

\end{document}